\documentclass{article}
\usepackage{graphicx} 
\usepackage{amsmath}
\usepackage{geometry}
\usepackage{caption}
\usepackage{lineno}

\usepackage{ragged2e}
\usepackage{setspace}
\usepackage{mathptmx}
\usepackage{url}
\usepackage{hyperref}
\newcommand{\myurl}[1]{{\urlstyle{same}\url{#1}}}
\usepackage{pgfplots}
\usepackage{booktabs}
\usepackage{tabularx}

\usepackage{tocbibind}
\usepackage{tocloft}

\newcommand{\nocontentsline}[3]{}
\newcommand{\tocless}[2]{\bgroup\let\addcontentsline=\nocontentsline#1{#2}\egroup}
\usepackage[T1]{fontenc}
\usepackage{authblk}
\usepackage{marvosym}
\usepackage[inkscapelatex=false]{svg}

\title{PAM: A Propagation-Based Model for Segmenting Any 3D Objects across Multi-Modal Medical Images}

\newcommand{\firstAuthorMark}{\textsuperscript{*}}
\newcommand{\correspondingAuthorMark}{\textsuperscript{\Letter}}

{\small  
\author{
    Zifan Chen\textsuperscript{1\firstAuthorMark},
    Xinyu Nan\textsuperscript{1\firstAuthorMark},  
    Jiazheng Li\textsuperscript{2\firstAuthorMark}, 
    Jie Zhao\textsuperscript{3}, 
    Haifeng Li\textsuperscript{4},  
    Ziling Lin\textsuperscript{1},
    Haoshen Li\textsuperscript{1},  
    Heyun Chen\textsuperscript{1},   
    Yiting Liu\textsuperscript{2},
    Lei Tang\textsuperscript{2\correspondingAuthorMark}, 
    Li Zhang\textsuperscript{1,4\correspondingAuthorMark}, and 
    Bin Dong\textsuperscript{4,5,6\correspondingAuthorMark} \\
    \textsuperscript{1}{Center for Data Science, Peking University, Beijing, China} \\
    \textsuperscript{2}{Department of Radiology, Key Laboratory of Carcinogenesis and Translational Research (Ministry of Education), Peking University Cancer Hospital and Institute, Beijing, China} \\
    \textsuperscript{3}{National Engineering Laboratory for Big Data Analysis and Applications, Peking University, Beijing, China} \\
    \textsuperscript{4}{Beijing International Center for Mathematical Research (BICMR), Peking University, Beijing, China} \\
    \textsuperscript{5}{Center for Machine Learning Research, Peking University, Beijing, China} \\
    \textsuperscript{6}{National Biomedical Imaging Center, Peking University, Beijing, China} 
}
}

\usepackage{pgfplots}
\begin{document}

\maketitle

\setstretch{1.5} 
\captionsetup{
  labelfont=bf,
  font={small, stretch=1.3}
}

\noindent {\firstAuthorMark These authors contributed equally:} Zifan Chen, Xinyu Nan, Jiazheng Li\\

\noindent \textbf{\correspondingAuthorMark Correspondence to:}
\begin{itemize}
\item Prof. Lei Tang, Department of Radiology, Key Laboratory of Carcinogenesis and Translational Research (Ministry of Education), Peking University Cancer Hospital and Institute, Haidian District, Beijing, 100142, China, tangl@bjcancer.org 
\item Dr. Li Zhang, Center for Data Science \& National Biomedical Imaging Center, Peking University, Haidian District, Beijing, 100080, China, zhangli\_pku@pku.edu.cn
\item Prof. Bin Dong, Beijing International Center for Mathematical Research (BICMR) \& Center for Machine Learning Research, Peking University \& National Biomedical Imaging Center, Peking University, Haidian District, Beijing,100080, China, dongbin@math.pku.edu.cn
\end{itemize}

\clearpage
\newpage
\section*{Abstract}
\noindent\textbf{Background:} Volumetric segmentation is crucial for medical imaging applications but faces significant challenges. Current approaches often require extensive manual annotations and scenario-specific model training, limiting their transferability across different tasks or modalities. While general segmentation models offer some versatility in natural image processing, they struggle with the unique characteristics of medical images. There is an urgent need in clinical practice for a new segmentation approach that can effectively handle medical imagery features while maintaining adaptability across various three-dimensional objects and imaging modalities.\\

\noindent\textbf{Methods:} We introduce PAM (Propagating Anything Model), a propagation-based segmentation approach that operates on 3D medical image volumes using a 2D prompt (bounding box or sketch mask). PAM extrapolates this initial input to generate a complete 3D segmentation by modeling inter-slice structural relationships, establishing a continuous information flow within 3D medical structures. This approach enhances segmentation effectiveness across various imaging modalities by focusing on structural and semantic continuities rather than isolating specific objects. The model combines a CNN-based UNet architecture for intra-slice information processing with a Transformer-based attention module to facilitate inter-slice propagation. This innovative framework results in a method with unique generalizability, capable of segmenting diverse 3D objects across different medical imaging modalities.\\

\noindent\textbf{Results:} PAM demonstrated superior performance on 44 diverse medical datasets, notably improving the dice similarity coefficient (DSC) for hundreds of segmentation object types and various medical imaging modalities. Compared to modern models like MedSAM and SegVol, PAM achieved an average DSC improvement of over 18.1\%, while maintaining stable predictions despite prompt deviation (one-way ANOVA test, $P\ge 0.5985$) and varying propagation configurations (one-way ANOVA test, $P\ge 0.6131$). Due to its efficient architecture and inference strategy, PAM exhibited significantly faster inference speeds (Wilcoxon rank-sum test, $P<0.001$) than existing models. The one-view prompt used by PAM also enhanced human prompt efficiency, reducing interaction time by about 63.6\% compared to the two-view prompts required by existing methods. Leveraging its focus on modeling inter-slice structural and semantic relationships, PAM demonstrated robust performance on unseen objects, addressing the challenge of limited transferability in current approaches. This generalizability was particularly evident in handling irregular and complex objects, with DSC improvements showing a negative correlation ($r<-0.1249$) to the degree of object irregularity. These results highlight PAM's ability to effectively capture the continuous flow of information within 3D medical structures. The source code and supplementary information are available on GitHub\footnote{https://github.com/czifan/PAM}. \\

\noindent\textbf{Conclusions:} PAM represents a significant advancement in medical image segmentation, offering a generalizable and versatile tool that addresses the challenges of limited transferability in current approaches. By efficiently generating accurate 3D segmentation from minimal 2D input across various imaging modalities, PAM demonstrates potential to reduce the need for extensive manual annotations and scenario-specific training in diverse medical contexts. This approach potentially paves the way for more automated and reliable medical imaging analyses in clinical applications.\\

\noindent\textbf{Key words:} Deep learning, segment any 3D objects, multiple modalities, medical images, volumetric segmentation, propagation-based model

\clearpage
\newpage

\section{Introduction}

Volumetric segmentation is a cornerstone task in medical image analysis~\cite{nnunet2021}, involving the precise identification and delineation of regions of interest (ROI) within three-dimensional (3D) medical images. This process is crucial for segmenting various anatomical structures such as organs, lesions, and tissues across a spectrum of imaging modalities, including computed tomography (CT), magnetic resonance imaging (MRI), positron emission tomography-computed tomography (PET-CT), and synchrotron radiation X-ray (SRX). The accurate segmentation of these structures is fundamental to a wide array of clinical applications, ranging from disease diagnosis~\cite{de2018clinically,yuan2023devil,cao2023large} and surgical planning~\cite{ferrari2012value,he2023associations} to monitoring disease progression~\cite{ouyang2020video,li2023ct,he2024deep} and optimizing therapeutic strategies~\cite{zaidi2010pet,bao2023deep,chen2024predicting}. However, 3D medical images present unique challenges for segmentation tasks, particularly due to the complex inter-slice relationships and the continuous nature of anatomical structures across slices, which are not encountered in traditional 2D image segmentation.

Despite advances in image analysis technology, manual segmentation remains the predominant method in many clinical scenarios~\cite{lu2021deep,he2024deep,li2023ct,chen2024predicting}. This process is not only time-consuming and labor-intensive but also requires high precision across diverse objects and imaging modalities~\cite{medsam2d2024}. These challenges underscore the pressing need for developing semi-automatic or fully automatic segmentation algorithms capable of handling any medical imaging modality and object. Such algorithms have the potential to significantly reduce the time and labor involved while improving the consistency of delineations~\cite{wang2018deepigeos,wang2021annotation}.

In response to these challenges, the past decade has witnessed significant advancements in deep learning-based models for medical image segmentation~\cite{unet2015,nnunet2021,dorent2023crossmoda}. These models have demonstrated remarkable capacity to learn complex image features and achieve precise segmentation across various tasks. However, a critical limitation of these approaches is their task-specific nature, often tailored to address particular segmentation challenges posed by specific medical imaging modalities and anatomical structures. This specialization necessitates the creation of large, meticulously annotated datasets for each new task, requiring medical experts to carefully delineate ROIs for specific objects and modalities~\cite{tang2020whole,primakov2022automated,xie2020relational,soomro2022image}. The process of data collection, manual annotation, and model training must be repeated for each new object or modality~\cite{nnunet2021}, an approach that is not only resource-intensive but also impractical for addressing emergent medical scenarios or rare pathologies. The substantial costs associated with data annotation and the scarcity of expert annotations exacerbate these challenges. Consequently, there is a growing demand for more generalized models capable of offering flexibility and rapid adaptability to new tasks without the need for repetitive, extensive training on narrowly defined datasets~\cite{antonelli2022medical}.

\begin{figure}[ht]
    \centering  
    \includegraphics[width=0.95\linewidth]{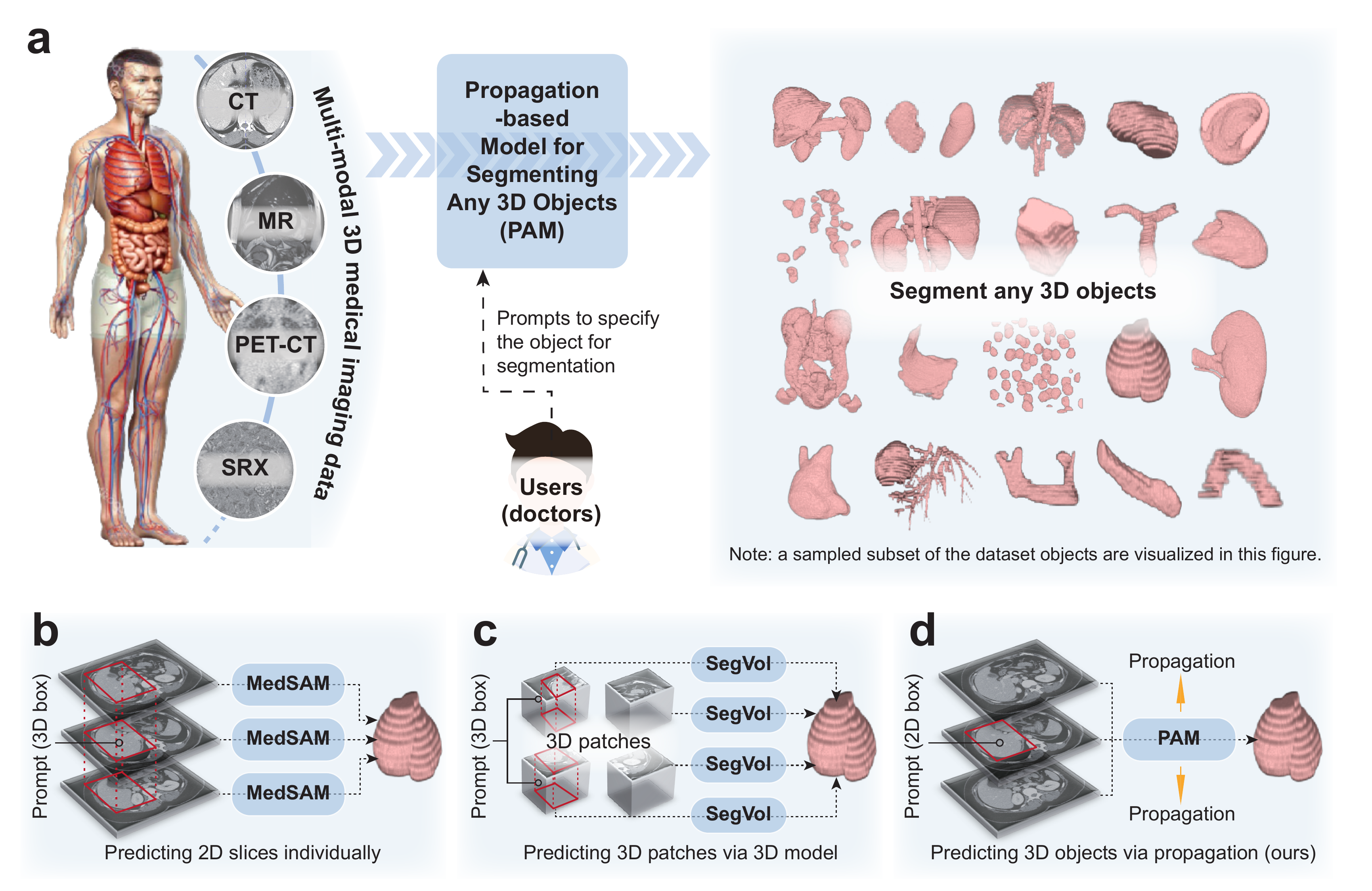}
    \caption{\textbf{PAM is designed for segmenting any 3D objects within various multi-modal 3D medical imaging data.} 
    \textbf{a} PAM receives any 3D medical imaging data as input, with users (typically doctors) specifying the target objects for segmentation through prompts. This enables precise and efficient volumetric segmentation of diverse 3D objects, thereby aiding users in enhancing the efficiency of medical analysis and diagnostics. 
    \textbf{b} Type I model: receives a 3D box prompt, predicts each 2D slice using a 2D model, and merges these 2D outcomes into a consolidated 3D prediction. 
    \textbf{c} Type II model: receives a 3D box prompt, predicts each 3D patch using a 3D model, and integrates these patch results into a comprehensive 3D prediction result.
    \textbf{d} Type III model (ours): receives a 2D box or mask prompt, employs a propagation model to disseminate prompt knowledge throughout across the entire 3D space, resulting in a unified 3D prediction.}
    \label{fig:overview}
\end{figure}

The advent of foundation models has revolutionized natural image processing, with the Segment Anything Model (SAM)~\cite{kirillov2023segment,ravi2024sam} exemplifying remarkable generalization capabilities across diverse tasks. This approach, leveraging user inputs such as points, bounding boxes, and masks, has proven highly effective for natural image segmentation. SAM's success is attributed to its training on vast and varied datasets, which enables it to establish robust correspondences between user prompts and segmentation results in natural images. Inspired by this breakthrough, researchers have begun adapting these versatile frameworks to the medical imaging domain, primarily through two types of models~\cite{medsam2d2024,du2023segvol,huang2024segment,mazurowski2023segment,deng2023segment,hu2023sam,he2023accuracy,wald2023sam,zhou2023can,wu2023medical,cheng2023sam}. 

The Type I model (Figure~\ref{fig:overview}b), often exemplified by MedSAM~\cite{medsam2d2024,zhu2024medical}, directly applies the SAM approach to various two-dimensional (2D) medical images. By training on a comprehensive collection of medical images, MedSAM can perform accurate object segmentation within 2D medical images based on simple user-provided 2D prompts. While promising in adapting SAM technology for medical use, Type I models struggle with the complexities of 3D medical imaging. Its lack of consideration for the continuity between adjacent slices in 3D image stacks results in significant challenges in achieving coherent volumetric segmentation. This limitation necessitates more complex user interactions, such as multiple prompts across different anatomical planes or dense annotations on each slice, to achieve satisfactory segmentation results. 

Recognizing these gaps, further research has led to the proposal of Type II models (Figure~\ref{fig:overview}c), such as SegVol~\cite{du2023segvol}, which aim to extend the SAM principles to 3D spaces by replacing 2D convolutional kernels with 3D counterparts. This approach enables Type II models to achieve smoother and more accurate 3D segmentation results compared to Type I models. Although Type II models have shown some success in processing volumetric data, they often struggle with generalizing to new, unseen objects or imaging modalities. Moreover, this approach introduces a massive number of parameters, substantially increases computational requirements, and leads to more complex user interactions, posing significant challenges for practical clinical applications.

Upon closer examination of the challenges faced by both Type I and Type II models, we observe that they essentially attempt to directly transplant SAM's approach of modeling correspondences between prompts and segmentations from natural images to 3D medical imaging. While this approach has proven effective for natural images, where objects often have clear boundaries or distinct semantic differences, it faces substantial limitations in medical imaging. Medical images typically exhibit subtle differences in pixel values and textures between objects, making it challenging to achieve generalizability through simple prompt-to-mask alignments. This realization highlights the pressing need to identify and leverage universal characteristics specific to medical images that can boost models' ability to generalize across diverse medical imaging scenarios. Building on this insight, we propose addressing these issues by introducing Type III models (Figure~\ref{fig:overview}d), which focus on modeling the continuous flow of information in 3D medical images. This concept of continuous flow of information, characterized by inter-slice relationships and the continuous nature of anatomical structures across slices, represents a fundamental distinction between 3D medical images and natural images. By explicitly modeling this characteristic, our approach not only resolves the issue of maintaining continuity when transitioning from 2D prompts to 3D segmentation but also leverages a universal property of medical images, thereby enhancing the model's generalization capabilities.

To implement this approach, we introduce PAM (Propagating Anything Model), an efficient framework to model the continuous flow of information within 3D medical structures. PAM achieves this through two main components: a bounding-box to mask module (Box2Mask), trained on over ten million medical images to respond to bounding-box-style prompts, and a propagation module (PropMask), trained on more than one million propagation tasks to model inter-slice relationships. The architecture employs convolutional neural networks (CNN) for local segmentation and a Transformer-based attention mechanism for modeling inter-slice information propagation. This hybrid design not only makes PAM more efficient than purely Transformer models (e.g., MedSAM and SegVol) in terms of parameters, computations, and inference speed, but also enables effective information propagation from a single 2D prompt to the entire 3D volume. As PAM's core task is to model structural and semantic relationships between slices, it exhibits generalizability across various medical imaging modalities and to new, unseen objects.

We have rigorously evaluated PAM through comprehensive experiments on 44 medical datasets, covering a variety of segmentation objects and medical imaging modalities. Experimental results demonstrate that PAM consistently outperforms the state-of-the-art (SOTA) segmentation foundation models with greater efficiency. Notably, PAM excels in handling irregular and complex anatomical structures, a common challenge in medical image segmentation. Its ability to capture and propagate intricate structural information allows for accurate delineation of objects with complex shapes, varying sizes, and inconsistent appearances across slices. This capability is particularly valuable in scenarios involving tumors or other anatomical structures with high variability. Moreover, PAM not only demonstrates robustness with unseen objects and maintains stability across deviated user prompts and different parameter configurations, but also quickly transforms into a powerful expert model for novel object types when fine-tuned with a small amount of annotated data. The fine-tuned PAM notably surpasses proprietary models that are trained from scratch on those limited annotated datasets. These results underscore the potential of PAM as a new paradigm for versatile volumetric medical image segmentation.

\begin{figure}[htp]
    \centering  
    \includegraphics[width=1.0\linewidth]{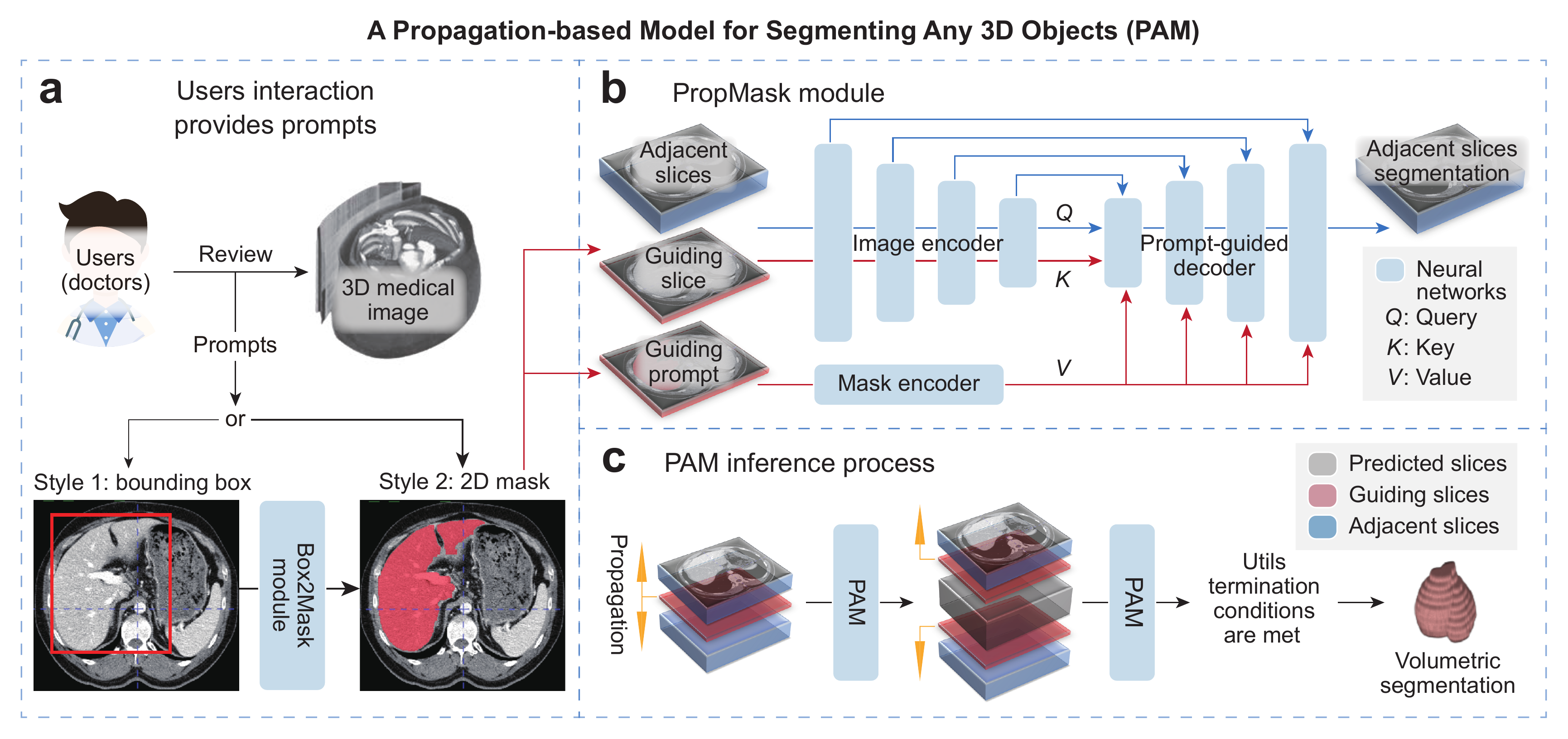}
    \caption{\textbf{Workflow and inference process of the propagation-based segment any 3D objects model (PAM).} 
    \textbf{a} User interaction: users upload a 3D medical image and specify the segmentation target using either a bounding box (style 1), in accordance with response evaluatin criteria in solid tumors (RECIST) guidelines, or a 2D mask applied to the largest slice of the target object (style 2). A bounding box is transformed into a 2D mask by the Box2Mask module for standardized processing in the PropMask module.
    \textbf{b} PropMask module: this module conducts volumetric segmentation by propagating information between slices. It begins with the 2D mask and its corresponding image slice (the guiding prompt and slice). Adjacent slices are the targets for segmentation. Image features (\textit{K} and \textit{Q}) are extracted from the guiding and adjacent slices, respectively, using a shared image encoder. The guiding prompt is converted into multi-scale features (\textit{V}) through a mask encoder. These features, along with skip connection features from adjacent slices, are assimilated in a prompt-guided decoder to facilitate volumetric segmentation, leveraging the propagation of prompt content across slices.
    \textbf{c} PAM inference: the user provides a guiding slice and prompt. PAM then propagates the prompt information bidirectionally across slices (yellow arrows). This propagation continues until the boundaries of the 3D image are reached or there is no further content to predict, achieving precise volumetric segmentation.
    }
    \label{fig:framework}
\end{figure}

\section{PAM: a Propagation-based Model for Volumetric Segmentation}
\label{subsec:propsam}


PAM focuses on learning the propagation of information across 2D slices in 3D medical images rather than on specific segmentation objects. As depicted in Figure~\ref{fig:framework}, the workflow of PAM begins with a user reviewing a 3D medical image and providing prompts within a slice for the target objects. PAM supports two types of prompts: 2D bounding boxes and sketch-based 2D masks (refer to Figure~\ref{fig:framework}a, Supplementary Text S1.1, and Supplementary Figure S2).

When a 2D bounding box is used, the Box2Mask module executes a foreground segmentation within the box, standardizing the input prompt format as sketch-based 2D masks for subsequent modules. Within PAM, the slice prompted by the user is termed the `guiding slice,' and the corresponding prompt is known as the `guiding prompt.' The PropMask module then plays a crucial role in utilizing the guiding prompt to segment adjacent slices of the same object (Figure~\ref{fig:framework}b). Initially, the guiding slice and adjacent slices are processed through a shared image encoder to extract image features, forming \textit{K} and \textit{Q}, respectively. Concurrently, the guiding prompt is transformed via a mask encoder into prompt-guided multi-scale features, termed \textit{V}. These features, along with multi-scale features extracted from adjacent slices (as skip connection features), are then combined in a prompt-guided decoder to predict volumetric segmentation. During this process, the PropMask module leverages the continuous flow of information between the guiding slice and adjacent slices to transfer the content of the guiding prompt to the adjacent slices, achieving effective volumetric segmentation.

During inference (Figure~\ref{fig:framework}c), PAM initially employs the user's prompt for a preliminary round of segmentation. Subsequent rounds utilize the most marginal slices from previous predictions as new guiding slices, enabling the propagation of the segmentation task through adjacent slices. This process continues iteratively until either the boundary of the 3D medical image is reached or PAM has no further content to predict.

\section{Results}

\subsection{Data characteristic and preprocessing}

\begin{figure}[htp]
    \centering  
    \includegraphics[width=0.95\linewidth]{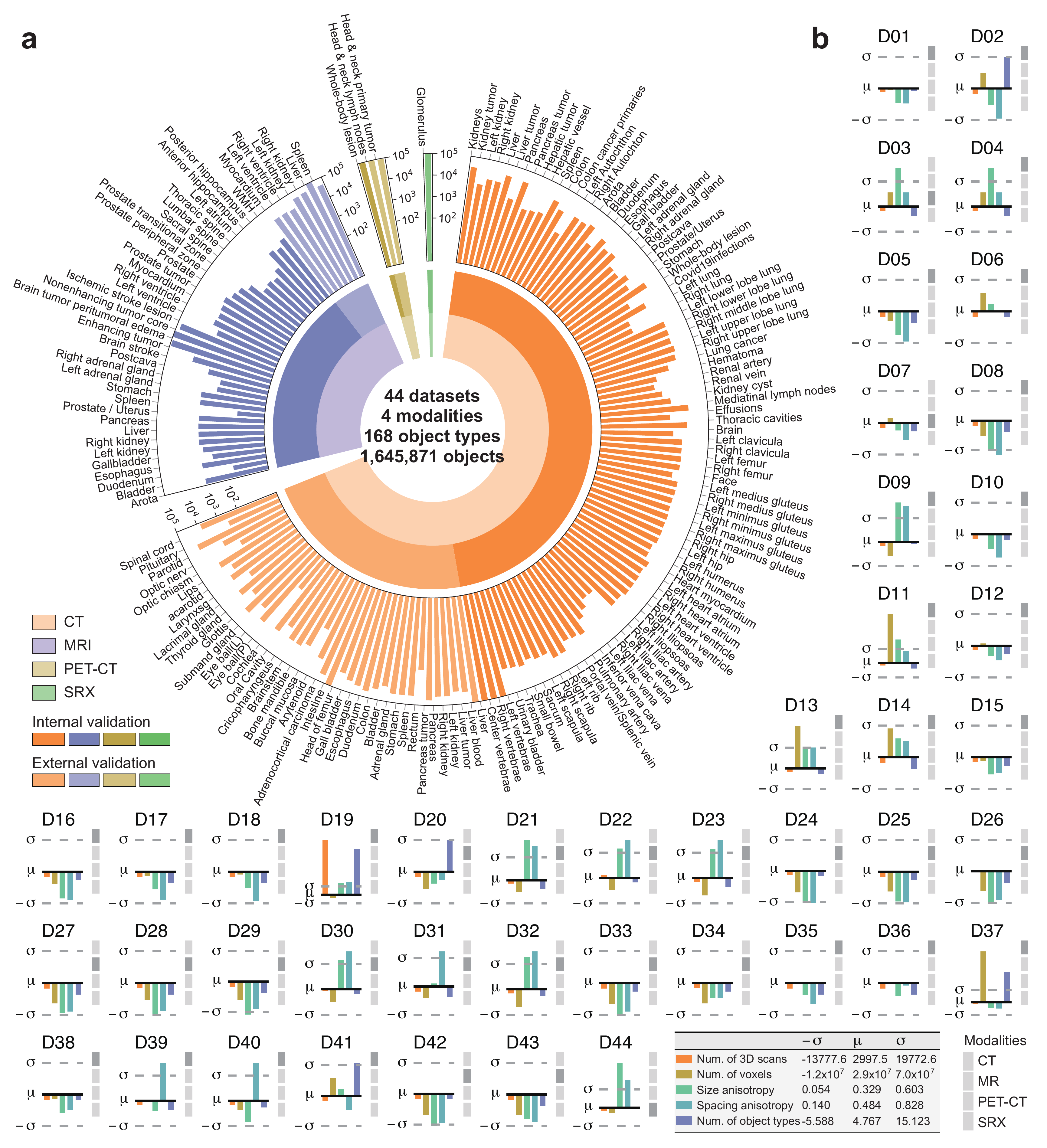}
    \caption{\textbf{Data characteristic across various datasets.} 
    \textbf{a} A circular barplot illustrates the range of data modalities and validation splits across multiple datasets. The innermost ring uses distinct colors to represent different medical imaging modalities (orange for CT; blue for MR; yellow for PET-CT; green for SRX). The second ring differentiates between internal and external datasets, with darker shades indicating internal datasets and lighter shades representing external datasets. The outermost layer displays a bar chart that showcases the distribution of segmented object types across the datasets, with quantities log-scaled for optimal visualization. 
    \textbf{b} Data fingerprints exhibit the key properties of the 44 datasets used in this study (displayed with z-score normalization over all datasets on a scale of one standard deviation around the mean). see Supplementary Tables S1–S4 for details.}
    \label{fig:data}
\end{figure}

This study utilized 44 publicly available 3D medical image datasets (Supplementary Table S1), encompassing various imaging modalities including CT, MRI, PET-CT, and SRX. These datasets cover 168 different target object types, totaling 1,645,871 3D objects for experimental analysis (Figure~\ref{fig:data}a). The diversity of these datasets is categorized across five dimensions (Supplementary Table S4): number of 3D scans, number of voxels, size anisotropy, spacing anisotropy, and variety of object types. This multidimensional diversity is crucial for a comprehensive evaluation of PAM, as illustrated in Figure~\ref{fig:data}b.

Size anisotropy, following nnUNet~\cite{nnunet2021}, is defined as the ratio of the smallest to the largest size in 3D scans, while spacing anisotropy is calculated as the ratio of the smallest to the largest spacing in 3D scans. In accordance with the protocol established in MedSAM~\cite{medsam2d2024}, we partitioned these datasets into 34 internal datasets (D01–D34) for training and validation, and ten external datasets (D35–D44) for independent testing (Supplementary Tables S2–S3).

As mentioned in Section~\ref{subsec:propsam}, PAM comprises two main modules: Box2Mask and PropMask. For training and evaluating the Box2Mask module, a 2D architecture model (detailed in Sections~\ref{subsec:data_acquisition}–\ref{subsec:data_preprocessing}), we processed 3D images and their 3D annotations through three steps. First, we simulated bounding boxes based on 3D masks to extract ROI images. Next, we normalized these ROI images. Finally, we applied random data augmentation to enhance the training data (details in Supplementary Text S1.3.2). Following this preprocessing, we obtained a total of 19,344,368 samples (2D medical image-mask pairs). These samples were divided into 14,974,620 training samples, 3,782,206 internal validation samples, and 587,542 external validation samples.

For training and evaluating the PropMask module, a 2D architecture model that receives both guiding slice and prompt and adjacent slices as inputs, we further preprocessed 3D images with their 3D annotations. This process involved determining the cropped size to extract both the guiding and adjacent slices, thereby constructing ROI tasks. We then normalized these ROI tasks and employed random data augmentation to enhance their training (further details in Supplementary Text S1.4.2). Following these preprocessing steps, we amassed a total of 1,345,871 tasks, each consisting of a guiding slice, a guiding prompt, and several adjacent slices. These tasks were distributed as follows: 1,020,576 for training, 258,889 for internal validation, and 66,406 for external validation.

\begin{figure}[htp]
    \centering  
    \includegraphics[width=1.0\linewidth]{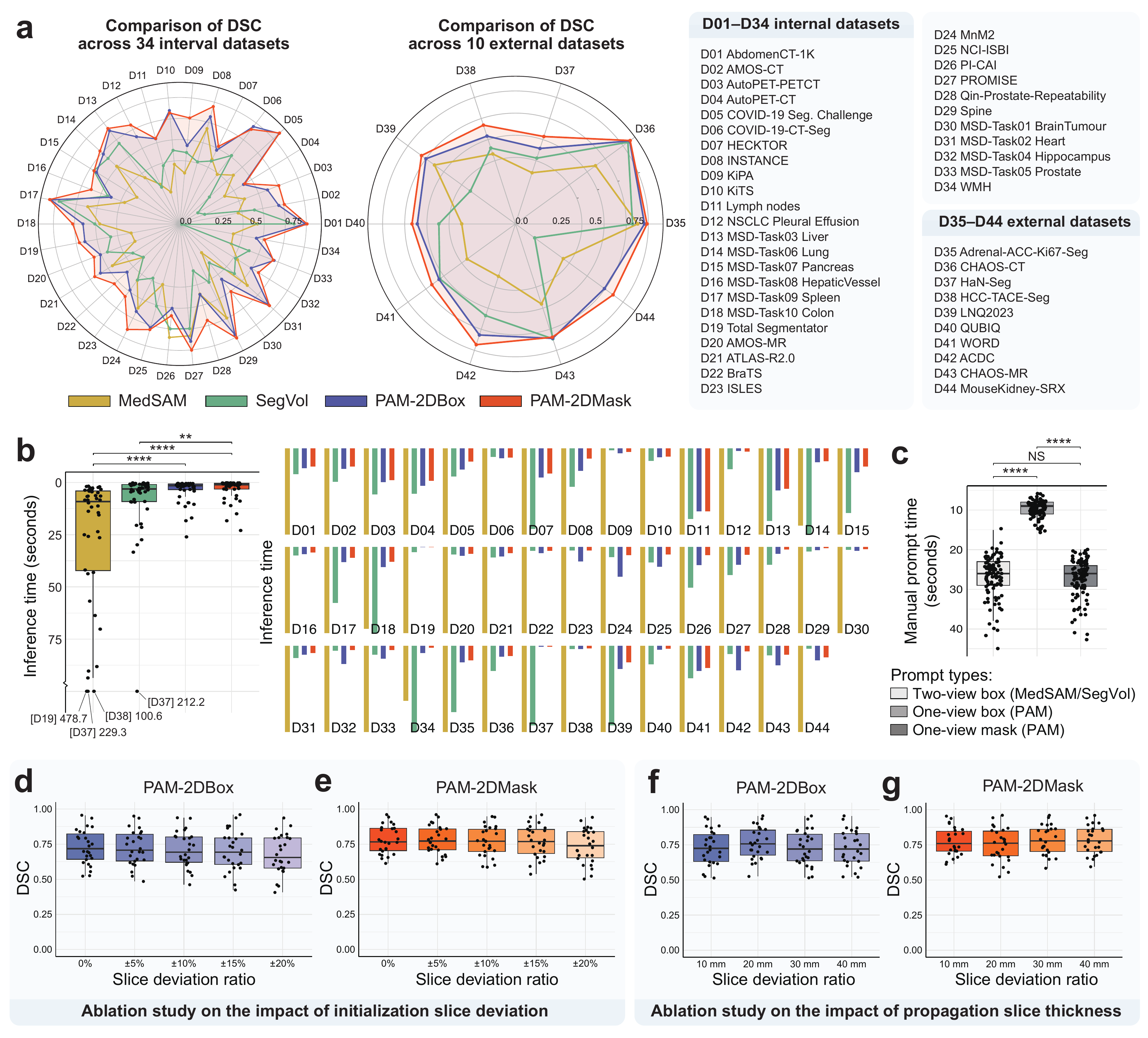}
    \caption{\textbf{Quantitative analysis of PAM across various datasets.}
    \textbf{a} Radar chart comparisons of Dice Similarity Coefficient (DSC) among four segmentation models—MedSAM (yellow), SegVol (green), PAM-2DBox (blue), and PAM-2DMask (red)—across internal and external datasets. Each radial axis represents one of the 44 datasets used (D1–D44), with DSC values ranging from 0.0 to 1.0, moving from the center outward.
    \textbf{b} Comparison of inference times (seconds). The left side features a box plot illustrating the distribution of inference times for the four models across 44 datasets. The right side visualizes a comparative analysis of the inference times for each model across these datasets, where the vertical axis represents inference time, and shorter bars indicate faster inference speeds.
    \textbf{c} Comparison of manual prompt times (seconds). A box plot depicts the distribution of interactive prompt times for three distinct prompt types. Light grey represents the commonly used two-view box prompt (MedSAM and SegVol), medium grey denotes the one-view box prompt of PAM, and dark grey signifies the one-view mask prompt of PAM. The vertical axis indicates the prompt times.
    \textbf{d–e} Ablation study on the impact of initialization slice deviation. Blue and orange colors represent the PAM-2DBox and PAM-2DMask models, respectively. The box plots show the distribution of DSCs plotted against initialization slice deviation from the RECIST-annotated maximum slice, with deviations of 0\% (no deviation), ±5\%, ±10\%, ±15\%, and ±20\%.
    \textbf{f–g} Ablation study on the impact of propagation slice thickness. The box plots display the distribution of DSCs plotted against propagation thickness of 10 mm, 20 mm, 30 mm, and 40 mm.
    }
    \label{fig:performance}
\end{figure}

\subsection{Segmentation Performance}

We evaluated two versions of PAM: PAM-2DBox, which accepts bounding-box-style (style 1) prompts, and PAM-2DMask, which receives mask-style (style 2) prompts. These were compared against two popular existing models, MedSAM and SegVol, on both internal and external datasets. Unlike PAM-2DBox, which requires only a single-view prompt such as a 2D bounding box, both MedSAM and SegVol necessitate two-view prompts (typically one bounding box on the axial plane and another on the orthogonal plane) within volumetric medical images. These two-view prompts form the tightest possible 3D bounding boxes of the segmentation targets, restricting the inference to the given slice area (Supplementary Figure S2).

MedSAM, originally designed for 2D medical image segmentation, requires processing each 2D medical slice containing the segmentation targets individually. The results from these segmentations are then stacked to form a volumetric final 3D segmentation, following guidance from its official GitHub\footnote{https://github.com/bowang-lab/MedSAM}. We refer to this process as `slice-by-slice prediction.' In contrast, SegVol directly segments volumetric medical images and employs a `zoom-out-zoom-in' strategy using resized global image and cropped patches as inputs to balance the acquisition of both global and local image features. We refer to this process as `patch-by-patch prediction.'

As illustrated in Figure~\ref{fig:performance}a, our two proposed PAMs exhibit superior segmentation performance (Supplementary Table S5), evaluated by the Dice Similarity Coefficient (DSC), across various experimental datasets. They achieve DSCs of 0.95 or higher on several segmentation objects (e.g., DSC=0.963 for livers, DSC=0.950 for kidneys, and DSC=0.950 for pancreas). Specifically, PAM-2DBox and PAM-2DMask both achieved higher DSCs on 31 of the 34 internal datasets and all ten external datasets compared to MedSAM and SegVol.

Overall, PAM-2DBox achieves an average DSC that is 21.0\% higher than MedSAM and 18.4\% higher than SegVol across all datasets. Similarly, PAM-2DMask demonstrates an average DSC that is 26.3\% higher than MedSAM and 23.7\% higher than SegVol. These results indicate the proposed PAMs' robust performance, while demonstrating outstanding performance on external validation datasets.

We observed that MedSAM does not demonstrate superior performance on any 3D segmentation tasks due to its `slice-by-slice prediction.' SegVol, while showing good performance on organ-related segmentation objects (e.g., DSC=0.941 for livers, DSC=0.912 for kidneys, and DSC=0.842 for pancreas), exhibits a notable decrease in performance on lesion-related or tissue-related segmentation objects (e.g., DSC=0.189 for whole-body lesions, DSC=0.001 for white matter hyperintensities, and DSC=0.161 for glomeruli). These limitations of SegVol stem from the general challenges of 3D segmentation models in limited medical image data, and the `patch-by-patch prediction' may cause fine information loss and discontinuity, posing challenges in predicting lesions with rare annotations and variable shapes.

In contrast, both PAM-2DBox and PAM-2DMask can accurately segment organ-related, lesion-related, and tissue-related segmentation objects. For example, their DSCs are 0.669 and 0.755 for whole-body lesions, 0.443 and 0.569 for white matter hyperintensities, and 0.888 and 0.886 for glomeruli, respectively. This indicates the exceptional generalization capabilities of PAM, stemming from its ability to learn generalized tasks (the continuous flow of information between slices rather than special objects).

These quantitative performance analyses underscore PAM's efficacy in accurately segmenting arbitrary 3D objects across a variety of medical imaging modalities and its potential for clinical applications.

\subsection{Inference and Interaction Efficiency}
\label{sec:res_efficiency}

We conducted a comprehensive evaluation of the inference times for PAMs, MedSAM, and SegVol across all datasets. As illustrated in Figure~\ref{fig:performance}b, MedSAM exhibits the slowest inference speeds (longest inference times), while SegVol shows an improvement over MedSAM. However, our proposed PAMs, in both the 2DBox and 2DMask versions, consistently achieve the fastest inference speeds (shortest inference times) (Wilcoxon rank-sum test, $P<0.001$). The right side of Figure~\ref{fig:performance}b visually details the specific inference time comparisons across the 44 datasets, with PAMs outperforming in nearly all cases (Supplementary Tables S6–S7).

The superior inference speed of PAMs can be attributed to its unique model structure and efficient inference strategy. PAM employs a hybrid architecture that combines a CNN-based structure, similar to UNet~\cite{unet2015}, with an attention mechanism inspired by Transformer architectures~\cite{transformer}. This approach allows PAM to leverage the strengths of both architectures for medical image segmentation while maintaining a relatively low parameter count. Specifically, the model has 32.48 M parameters, which increases to 53.1 M parameters when combined with the Box2Mask module for supporting bounding-box prompts.

The inference strategies of the compared models showcase significant differences in their approaches to 3D medical image segmentation. MedSAM, a type I model, processes each slice individually using a complex Transformer model (Figure~\ref{fig:overview}b). SegVol, classified as a type II model, adopts an inference process similar to the 3D-nnUNet model, predicting individual patches with dense overlapping strides that are later merged (Figure~\ref{fig:overview}c). In contrast, PAM, a type III model, utilizes a 2D model structure and performs bidirectional parallel inference without the need for overlapping window slides (Figure~\ref{fig:overview}d).

Quantitative analysis of computational resources and inference times across the 44 datasets reveals that PAM consistently requires less computational resources and achieves faster inference times compared to both MedSAM and SegVol (Figure~\ref{fig:performance}b and Supplementary Tables S6–S7). These results suggest that PAM's architectural design and inference strategy contribute to its efficiency in segmenting 3D medical images, potentially offering advantages in clinical applications where rapid and accurate segmentation is crucial.

\begin{figure}[htp]
    \centering  
    \includegraphics[width=1.0\linewidth]{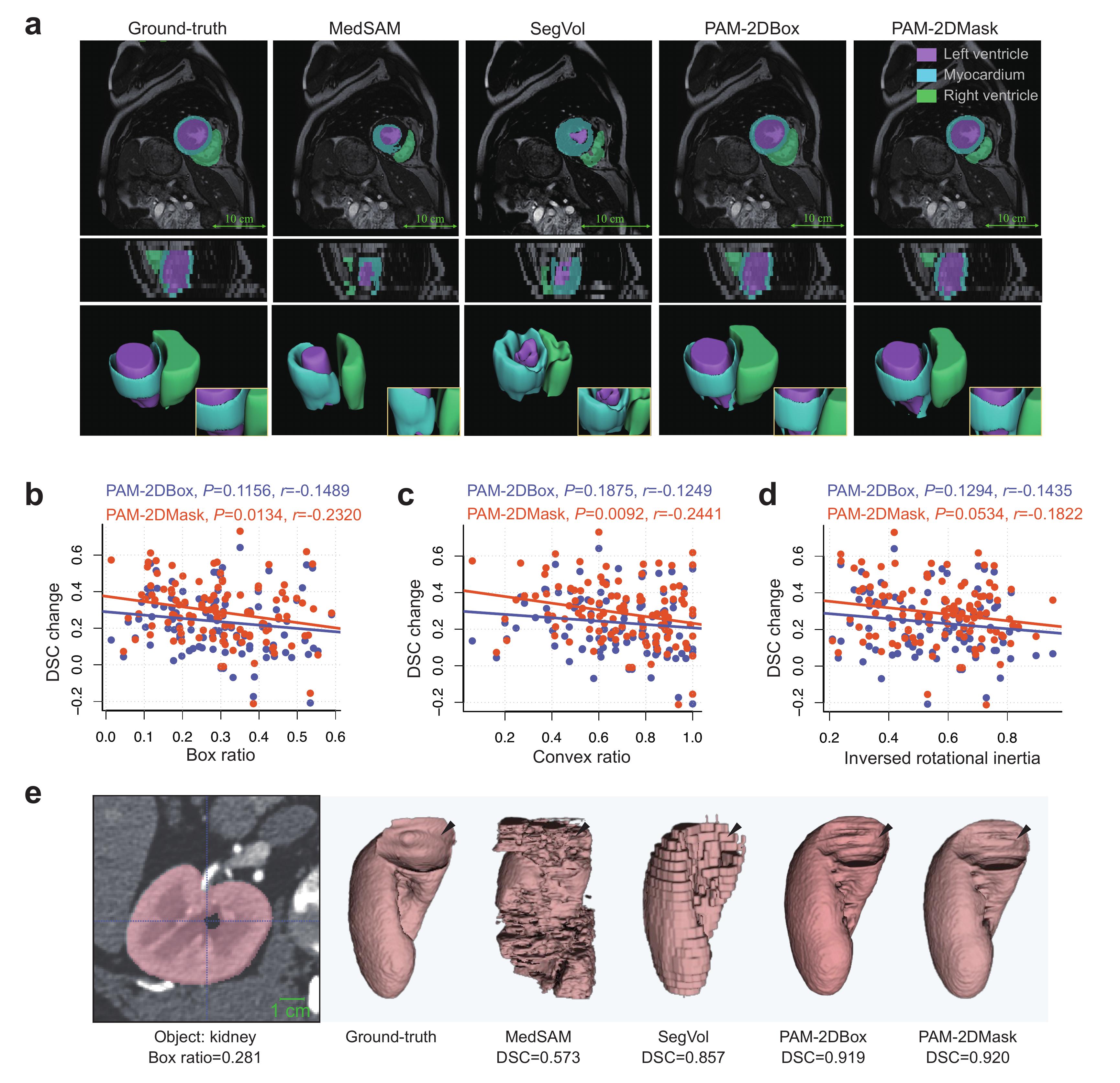}
    \caption{\textbf{Qualitative analysis and the relationship between object shape and performance.}
    \textbf{a} Comparison of segmentation results across various models. From left to right, the columns represent ground truth, MedSAM, SegVol, PAM-2DBox, and PAM-2DMask, respectively.
    \textbf{b} DSC change analysis for PAM relative to MedSAM across various box ratios. This plot displays DSC changes, where blue indicates PAM-2DBox and red denotes PAM-2DMask. Each point represents a different object, highlighting the model's adaptability to varying box ratios.
    \textbf{c} DSC change analysis for PAM relative to MedSAM across various convex ratios.
    \textbf{d} DSC change analysis for PAM relative to MedSAM across various inverse rotational inertia (IRI).
    \textbf{e} Comparative visualization of segmentation results for a sample with a low box ratio. The black arrow indicates an irregular area.}
    \label{fig:convex analysis}
\end{figure}

We also explored the interaction efficiency of different models. Both MedSAM and SegVol require two-view prompts, whereas PAM-2DBox only necessitates interaction in one view. In our extracted test subset (see Supplementary Text S2 for experimental details), an experienced radiologist interacted with different datasets and objects using various interaction prompts. We recorded the time taken for each interaction and compared the different prompt types.

As demonstrated in Figure~\ref{fig:performance}c, the one-view box prompt of PAM took significantly less time than the common two-view box prompt used in MedSAM and SegVol (Wilcoxon rank-sum test, $P<0.0001$), reducing interaction time by approximately 63.6\% and aligning closely with the practical needs of clinical practitioners. Although the one-view mask prompt (used in PAM-2DMask) requires more time than one-view box prompts, it offers a comparable interactive cost to the two-view box prompts ($P=0.3063$), and the more detailed prompt information can greatly enhance the model's overall performance (Figure~\ref{fig:performance}a). These analyses demonstrate the advantages of PAM's two types of prompts over the two-view box prompt, providing users with flexible options for practical application.

\subsection{Stability and Consistency}

PAM operates with one-view prompts, typically selected by physicians according to Response Evaluation Criteria in Solid Tumors (RECIST). To assess the impact of variations in prompts provided by different physicians on PAM's performance, we conducted an ablation study. As depicted in Figure~\ref{fig:performance}d–e, we simulated deviations from the RECIST-standard confirmed largest slice through five experimental groups: 0\% (no deviation), ±5\%, ±10\%, ±15\%, and ±20\%.

Both PAM-2DBox and PAM-2DMask demonstrated stable DSC across these variations, as confirmed by one-way ANOVA tests, with $P$-values of 0.6736 and 0.5985, respectively (see Supplementary Figure S9 and Supplementary Table S8 for further details). While performance slightly declined with increasing deviations, it is noteworthy that a deviation of ±20\%, which corresponds to a total range of 40\%, is uncommon in clinical practice. Even with such substantial deviations, PAMs maintained commendable performance.

Moreover, during the inference process, PAMs iteratively select the most marginal predicted slice as the next round's guiding slice and guiding prompt. The distance of this slice from the original guiding slice could potentially affect the accuracy of subsequent predictions, particularly when far apart, as the propagation relationship weakens with distance. This influence may be amplified through iterative propagation, impacting the overall 3D segmentation of the object.

To evaluate the impact of propagation slice thickness, we conducted another ablation study with propagation thicknesses of 10 mm, 20 mm, 30 mm, and 40 mm. As shown in Figure~\ref{fig:performance}f–g, PAMs maintained predictive stability and consistency across these different thicknesses, as assessed by one-way ANOVA tests with $P$-values of 0.7114 and 0.6131, respectively (refer to Supplementary Figure S10 and Supplementary Table S9 for more details). Based on these findings, we empirically selected 20 mm as the default propagation thickness for PAMs. 

Through these ablation experiments involving varied prompt deviations and propagation thicknesses, PAMs have demonstrated notable predictive stability and consistency. These results provide a reliable foundation for the potential clinical application of PAMs.

\begin{figure}[htp]
    \centering  
    \includegraphics[width=1.0\linewidth]{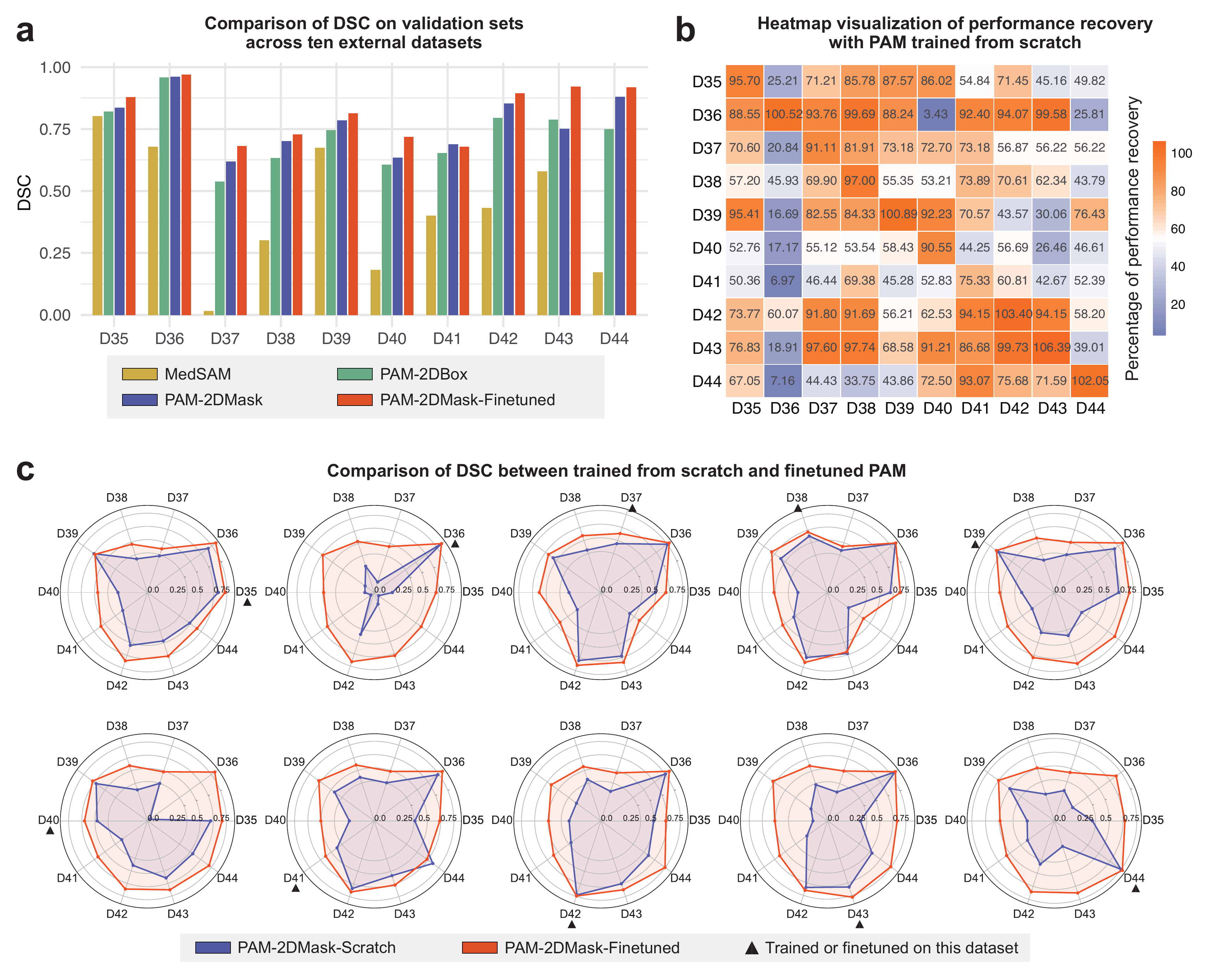}
    \caption{\textbf{Generalization analysis of PAM across multiple datasets.}
    \textbf{a} Bar chart visualization of DSC across external datasets. This bar chart compares the DSC performance of four models on validation sets from ten external datasets. The models compared inlcude MedSAM in yellow, PAM-2DBox in green, PAM-2DMask in blue, and PAM-2DMask finetuned on corresponding datasets in red. The horizontal axis labels datasets D35 to D44, and the vertical axis represents the DSC performance.
    \textbf{b} Heatmap of performance recovery. The heatmap displays the percentage of performance recovery for PAM trained from scratch across ten datasets (each column) compared to the original trained PAM. Red indicates a recovery of more than 50\%, with deeper reds showing higher percentages. Blue indicates recoveries below 50\%, with deeper blues denoting lower percentages.
    \textbf{c} Radar chart comparison of DSC between trained from scratch and finetuned PAM. This radar chart visualizes the DSC comparison between PAM trained from scratch and finetuned on original trained PAM across datasets D35 to D44. Blue lines represent PAM trained from scratch, while red indicated finetuned PAM. Triangular markers point to the datasets used for training or finetuning.}
    \label{fig:robustness}
\end{figure}

\subsection{Efficacy in Segmenting Complex and Irregular Objects}

We visualized the qualitative segmentation results of different models as shown in Figure~\ref{fig:convex analysis}a. Our proposed PAMs effectively utilize propagation information between slices, resulting in visually complete and smooth segmentation outcomes. In contrast, MedSAM, which employs a `slice-by-slice prediction' and merging strategy, and SegVol, which uses `patch-by-patch prediction' and integration, do not achieve segmentation visualizations as refined as those produced by PAMs.

We observed that segmentation difficulty varies among different objects. For instance, most organs have relatively fixed shapes, making them easier to learn and segment, whereas some tissues or lesion-related objects present greater challenges. To quantify this process, inspired by BiomedParse~\cite{zhao2024biomedparsebiomedicalfoundationmodel}, we evaluated the `irregularity' of the objects and the accuracy of our predictions for these irregular objects through three metrics: Box ratio, Convex ratio, and Inverse rotational inertia (IRI), as defined in Section~\ref{sec:object_complexity_metrics}.

As demonstrated in Figure~\ref{fig:convex analysis}b–d, PAM shows performance improvements over MedSAM across most segmentation objects (DSC change > 0). Furthermore, these improvements are more pronounced when the three irregularity metrics are smaller ($r<-0.1249$), indicating that PAM is particularly effective at handling irregular objects and more accurately reflects real-world challenges.

Additionally, we visualized the segmentation results for four models on a sample with a box ratio of 0.281 in Figure~\ref{fig:convex analysis}e. The qualitative results of PAM are clearly closest to the ground truth. This is notable especially since the segmentation object in this sample, unlike a typical kidney, includes a tumor, resulting in an irregularly indented shape indicated by the black arrow. PAM successfully identifies this unique feature, whereas MedSAM and SegVol struggle to accurately discern this specific part.

This success underscores that PAM's learning focus is more robust, enabling it to dynamically adapt to rare and complex shape variations based on changes in structure and semantics between slices. This ability aligns with our initial premise of modeling the continuous flow of information within 3D medical structures. In contrast, MedSAM and SegVol may find these abnormalities challenging due to their reliance on features typical of a normal kidney, highlighting the limitations of approaches that do not explicitly account for inter-slice relationships. These analyses demonstrate the potential of PAMs for precise segmentation of various objects, particularly those with irregular shapes.

\subsection{Generalization and Adaptability Across Diverse Segmentation Tasks}

We explored PAM's generalization capabilities from two perspectives: model fine-tuning and training from scratch. As depicted in Figure~\ref{fig:robustness}a, PAM, serving as a general segmentation model, outperforms MedSAM on ten external datasets, demonstrating its robust generalization ability to unfamiliar datasets and objects. We then partitioned these ten datasets into training and validation sets and conducted minimal fine-tuning of PAM on the training sets to create the PAM-2DMask-Finetuned model. Experiments indicate that with minimal data fine-tuning, PAMs can quickly adapt to corresponding tasks and enhance performance. However, the improvement of PAM-2DMask-Finetuned over PAM-2DMask is less significant compared to PAM's improvement over MedSAM (paired t-test, $P=0.0011$), suggesting that the general model of PAM already performs well on unseen objects.

We also trained PAM from scratch on the divided training set and evaluated the performance recovery percentage relative to the general model PAM across ten datasets (as shown in Figure~\ref{fig:robustness}b). The experiments reveal that even with limited data, training PAM from scratch can achieve over 75.33\% performance recovery on corresponding datasets, indicating that PAM's learning tasks are sufficiently straightforward to allow rapid adaptation on limited samples. Furthermore, we observed that segmentation objects with similar structures to the training objects benefited in performance. For instance, when trained from scratch on dataset D35, the performance recovery on D39 reached 87.57\% due to the objects in these two datasets being both lesion-related and structurally similar. This further underscores that PAM's learning focus is not on specific semantic objects, but rather on the structural or semantic information transfer relationships between slices, aligning with our aim to model the continuous flow of information within 3D medical structures.

Using radar charts (Figure~\ref{fig:robustness}c), we showcased the segmentation performance on ten datasets using the general model fine-tuned (PAM-2DMask-Finetuned) and trained from scratch (PAM-2DMask-Scratch). As noted, aside from performing well on the datasets where it was fine-tuned, PAM-2DMask-Scratch also achieved commendable performance on datasets with similar structural segmentation objects. Generally, the performance of PAM-2Dmask-Finetuned is superior to PAM-2DMask-Scratch, indicating that the general capabilities aid in fine-tuning specific objects and achieving more precise segmentation results~\cite{xie2024sam,li2024polyp}. This also demonstrates PAM's adequate general capability to handle various segmentation objects, reinforcing its effectiveness in capturing inter-slice relationships across diverse medical imaging scenarios.

These analyses highlight PAM's robust generalization capabilities, showcasing its effectiveness in both fine-tuned and from-scratch training scenarios across diverse datasets and structural variations. The results affirm PAM's ability to model the continuous flow of information within 3D medical structures, enabling adaptability and precision in segmentation tasks across various imaging modalities and object types.

\section{Discussion}


We developed PAM (Propagating Anything Model) to address the critical need for efficient and accurate volumetric segmentation across diverse 3D medical imaging modalities. PAM offers several key advantages over existing approaches. Using only a single 2D prompt, it achieves high-quality segmentation of any 3D object across a wide range of medical imaging tasks, significantly outperforming existing state-of-the-art methods without being constrained by predefined object categories or specific modalities. Unlike traditional methods, PAM models the continuous flow of information in 3D structures through an innovative propagation framework, capturing inter-slice relationships and going beyond simple object-specific feature learning. This approach enables PAM to demonstrate superior performance and generalization capabilities, making it a versatile tool for various 3D medical segmentation challenges.

The application of successful natural image processing approaches like SAM to medical imaging often struggles with unique challenges posed by 3D medical data, resulting in performance degradation and limited generalization. This gap can be attributed to several factors, primarily the limited availability of annotated data in medical imaging compared to natural image datasets, and the semantic ambiguity in medical objects, particularly in pathological structures. SAM-like methods typically focus on learning to segment specific objects based on prompts. While this approach is effective for 2D natural images due to the abundance of diverse samples, it faces significant challenges in 3D medical imaging. The combination of limited data and high semantic complexity in medical volumes often causes these models to overfit to a small set of object patterns present in the training data, limiting their ability to generalize to the wide variety of structures and anomalies encountered in clinical practice.

In contrast, PAM addresses these challenges by modeling the continuous flow of information across slices, a unique characteristic of 3D medical structures. This approach allows PAM to learn generalizable inter-slice relationships rather than relying on object-specific features. Consequently, PAM avoids the trap of overfitting to limited object patterns and instead captures the underlying structural and semantic continuity across slices. This novel approach enables PAM to better adapt to the diverse and complex nature of 3D medical objects, demonstrating superior generalization to unseen objects and new imaging modalities, even with limited training data.


At the core of PAM is a novel propagation-based segmentation model that integrates CNN-based local feature extraction with Transformer-based attention mechanisms for modeling long-range dependencies. This hybrid architecture enables PAM to efficiently extract precise local features crucial for accurate medical image segmentation while effectively modeling inter-slice relationships and propagating information throughout the volume. As a result, PAM achieves a balance between model capacity and computational efficiency, resulting in fewer parameters compared to purely Transformer-based models. This approach contrasts with other methods that either apply 2D segmentation slice-by-slice (e.g., MedSAM) or attempt to model 3D structures directly at the cost of increased computational demands and limited generalization (e.g., SegVol).

Our experimental results demonstrate PAM's superior performance across a wide range of datasets and imaging modalities. By focusing on learning inter-slice relationships, PAM excels at capturing and propagating intricate structural information, enabling it to effectively segment irregular objects. This capability is particularly valuable for complex anatomical structures and pathological features that deviate from typical shapes. The model's effectiveness in handling such challenging cases underscores its potential to significantly improve medical image analysis in various clinical applications.

Despite these promising results, achieving general volumetric segmentation in medical imaging remains challenging. The diversity of objects and modalities in medical imaging presents a significant hurdle, with variability in object shapes, sizes, and contrasts across different imaging techniques. Additionally, the limited availability of large-scale, annotated 3D medical imaging datasets poses challenges for training and evaluation. These factors highlight the importance of developing robust and adaptable models like PAM that can generalize across diverse medical imaging scenarios.

To address these ongoing challenges and further improve PAM's capabilities, we propose several directions for future work. First, evaluating PAM's impact on downstream clinical tasks by integrating it into clinical workflows and assessing its practical implications for patient diagnostics and prognosis will be crucial. Exploring additional interactive input methods, such as point-based or line-based prompts, could enhance flexibility and user experience. Investigating mechanisms to capture and utilize global contextual information more effectively may improve the segmentation of large or discontinuous structures. Furthermore, exploring the integration of diverse imaging modalities and complementary data types could enhance segmentation accuracy and robustness. Finally, developing strategies for rapid adaptation to new imaging modalities or specific object types with minimal additional training data will be essential for PAM's widespread adoption in varied clinical settings.


In conclusion, PAM represents a significant advancement in 3D medical image segmentation, demonstrating exceptional segmentation capabilities and strong generalization abilities across 44 datasets and multiple medical imaging modalities. By focusing on modeling the continuous flow of information within 3D medical structures, PAM not only achieves superior performance, particularly for complex and irregular objects, but also offers higher inference and interaction efficiencies. These advantages position PAM as a versatile and efficient solution for volumetric segmentation across diverse imaging modalities. Moreover, PAM's success in leveraging the continuous flow of information as a unique learning target opens up new avenues for future research in 3D medical image segmentation. We encourage future research to explore similar innovative learning objectives that capitalize on the inherent characteristics of medical imaging data. Such approaches could include modeling temporal dynamics in 4D imaging, exploring inter-modality information flow, or investigating hierarchical spatial relationships within complex anatomical structures. By identifying and leveraging these unique aspects of medical imaging data, future models may achieve greater performance and generalization capabilities, further advancing the field of medical image segmentation.

\section{Methods}
\subsection{Data acquisition}
\label{subsec:data_acquisition}
We collected 44 public 3D medical segmentation datasets (see Supplementary Table S1) encompassing multiple modalities, including CT, MR, PET-CT, and SRX, to construct a large-scale and comprehensive dataset for model training and validation (Figure~\ref{fig:overview}a). These datasets have been widely used for training and validating both universal and specialized medical image segmentation models, ensuring that all volumetric medical images in our dataset possess high-quality annotations.

As illustrated in Supplementary Table S4 and Figure~\ref{fig:data}b, we characterized these 44 datasets across five dimensions: number of 3D scans, number of voxels, size anisotropy, spacing anisotropy, and variety of object types. These dimensions are essential for a comprehensive evaluation of PAMs. Following nnUNet~\cite{nnunet2021}, we defined size anisotropy as the ratio of the smallest to the largest size in 3D scans, and spacing anisotropy as the ratio of the smallest to the largest spacing in 3D scans. Furthermore, in alignment with the protocol established in MedSAM~\cite{medsam2d2024}, we partitioned these datasets into 34 internal datasets (D01–D34) for training and validation, and ten external datasets (D35–D44) for independent testing (Supplementary Tables S2–S3).

As shown in Supplementary Tables S2 and S4, our dataset comprises 168 different categories of objects and 1,645,871 3D objects, covering organs, lesions, and tissues. This diverse collection enhances the stability and generalization capability of PAM.

\subsection{Data pre-processing}
\label{subsec:data_preprocessing}
Our data pre-processing pipeline consists of several steps to prepare the data for the Box2Mask module and the PropMask module.

\subsubsection{Bounding Box Generation for Box2Mask}
To obtain bounding boxes for the Box2Mask module, we generated the tightest bounding box on the slice where the corresponding foreground mask annotation contains over 100 pixels. We then randomly adjusted the width and height of the bounding box with a scaling ratio between 1.0 to 1.25 to account for potential deviation in actual usage. These processed bounding boxes were used as training data for the Box2Mask module.

\subsubsection{ROI Task Construction for PropMask}
For the PropMask module, we constructed ROI tasks. We generated the tightest bounding box around the mask of the guiding slice and then randomly adjusted its width and height with a scaling ratio between 1.0 to 2.0 to capture the context around the target object. This adjusted bounding box was then used to crop both the guiding slice and the adjacent slices sampled within the propagation thickness, forming the cropped ROI tasks as training data for the PropMask module.

\subsubsection{Image Normalization}
After acquiring the ROI images/tasks, we normalized the intensity values by clipping them to the range between the 0.5th and 99.5th percentiles of pixel values within the annotated mask of the original slice images. This process enhanced and emphasized the context of foreground ROI images/tasks.

\subsubsection{Data Augmentation}
To optimize training efficiency, we applied offline data augmentation for both the Box2Mask and PropMask modules. For the Box2Mask module, we augmented each sample five times. Each image had a 50\% chance of being flipped horizontally and vertically. Additionally, we randomly adjusted the image's brightness and contrast, also with a 50\% probability, setting the adjustment ranges to [-0.2, 0.2]. The images were also rotated randomly up to 45 degrees with a 50\% probability, filling any areas outside the original boundaries with a constant value (typically black).

For the PropMask module, since the fundamental training unit is a task containing several images (typically 20 adjacent images and one guiding image), we applied augmentation to each image within a task. Specifically, each image in a task had a 50\% chance of being flipped horizontally or vertically and being rotated up to 45 degrees.

After augmentation, all samples were uniformly resized to a resolution of $224\time224$ for input into both the Box2Mask and PropMask modules.

\subsubsection{Dataset Partitioning}
Following these pre-processing steps, we obtained a total of 19,344,368 samples for the Box2Mask module and 1,345,871 tasks across 44 datasets. In accordance with MedSAM's data partitioning protocol, we divided these data into internal and external validation datasets. The internal validation dataset was further split into training and validation sets at an 80:20 ratio.
The final distribution of samples was as follows:
\begin{itemize}
    \item Box2Mask module: 14,974,620 training samples, 3,782,206 internal validation samples, and 587,542 external validation samples (Supplementary Table S2).
    \item PropMask module: 1,020,576 training samples, 258,889 internal validation tasks, and 66,406 external validation tasks (Supplementary Table S3).
\end{itemize}

We trained the overall PAM on the training set, using the internal validation set to evaluate model performance and select the final model checkpoint. The external validation dataset served to demonstrate the robustness of PAM and its zero-shot capability with unseen objects and datasets.

\subsection{Network architecture}

PAM consists of two main components: the Box2Mask module and the PropMask module. Both are built on convolutional neural networks (CNNs), which have long been the dominant architecture in computer vision. CNNs offer greater efficiency compared to the currently popular Transformer architecture, making them suitable for a wider range of clinical applications. We note that the UNet-based architecture\cite{unet2015}, particularly the nnUNet model\cite{nnunet2021}, has become the most widely adopted and effective approach for medical image segmentation in recent years.

\subsubsection{Box2Mask Module}
The Box2Mask module is designed to convert ROI images, cropped according to bounding box prompts, into binary foreground masks (Figure~\ref{fig:framework}a). We employed a six-stage encoder-decoder UNet-based network\cite{unet2015} for this purpose. The input utilizes three-dimensional channels, suitable for a grayscale image replicated three times. The network structure is as follows:
\begin{itemize}
    \item Initial stage features 32 channels, doubling with each subsequent stage up to 512.
    \item Channel counts across the six stages: [32, 64, 128, 256, 512, 512].
    \item Each stage includes two convolutional layers, followed by instance normalization\cite{instancenorm2016} and LeakyReLU activation.
    \item All convolutional kernels are 3x3, with a stride of one within each stage and a stride of two in the last layer of each stage for down-sampling.
    \item Skip connections bridge each encoder level with its corresponding decoder to maintain low-frequency features.
    \item Binary foreground segmentation predictions are produced at all six decoding stages for deep supervision.
\end{itemize}

\subsubsection{PropMask Module}
The PropMask module uses a 2D mask as input for the initial guiding slice. This 2D mask can either be generated by the Box2Mask module or provided directly by the user. It then propagates this segmentation to adjacent slices (Figure~\ref{fig:framework}b). The architecture of PropMask, which is the core component of the network, is largely based on UNet\cite{unet2015} and consists of: an image encoder, a mask encoder, a sequence of cross-attention modules, and a decoder.

Both the image encoder and mask encoder are six-stage CNN encoders, similar to the Box2Mask encoder. However, the mask encoder's input channel is one to directly accept the 2D mask prompt. The encoders process the guiding slice, adjacent slices, and the 2D mask to produce features at six resolutions: $([224\times224, 112\times112, 56\times56, 28\times28, 14\times14, 7\times7])$. Similarly, the 2D mask from the guiding slice go through the mask encoder to produce mask features of six resolutions. 

Cross-attention modules are then employed, defined as:
\begin{align*}
    &\text{Attention}(Q, K, V) = \text{Softmax}\left(\frac{QK^T}{\sqrt{d_k}}\right)V 
\end{align*}
where Q, K, and V are query, key, and value vectors respectively. $QK^T$ represents the dot product between queries and keys, measuring the similarity or alignment between them. The scaling factor $\sqrt{d_k}$ helps maintain stable gradients during training. Cross-attention is particularly useful when the sets of queries, keys, and values are derived from different input sources, enabling the model to integrate information across these sources. In our case, it allows the PropMask module to effectively combine information from the guiding slice, adjacent slices, and the input mask.

In the PropMask module, support features, query features, and mask features are flattened into 1-dimensional vectors, serving as the support, query, and value vectors for cross-attention respectively. The outputs of the cross-attention modules can be regarded as the value vectors for the query features. These output value vectors are then reshaped into 2D feature maps, which serve as the feature maps for query images.

To balance model efficiency and performance, cross-attention modules are applied only to the lowest four resolution feature maps: $([56\times56, 28\times28, 14\times14, 7\times7])$. The output from the lowest resolution cross-attention module goes through multiple de-convolutions in the decoder, generating outputs that match the shapes of the six-stage encoder's image features.

Following the UNet architecture's skip connection structure\cite{unet2015}, each decoder stage's output is concatenated with feature maps of the same resolution from either the cross-attention module or the encoder stage. The final decoder output is the prediction mask, which is used to calculate dice loss with the ground truth mask label for backpropagation.

\subsection{Training configuration and inference settings}
We implemented our models using PyTorch\cite{paszke2019pytorch} (version 2.0.0) and executed them on a server equipped with the CUDA platform (version 11.8). Both the Box2Mask module and the PropMask module were trained using four NVIDIA A800-SXM4-80GB GPUs and 64 Intel(R) Xeon(R) Platinum 8358 P CPUs (2.60GHz). We utilized the AdamW optimizer with an initial learning rate of 1e-3 for the Box2Mask module and 5e-4 for the PropMask module, as well as a weight decay of 1e-4. The learning rate was adjusted according to the Cosine Annealing LR schedule with a maximum period of 100 epochs and a minimum eta of 1e-5.

For the Box2Mask module, during each epoch, we randomly selected 10,000 samples for training and conducted evaluations every 20 epochs using a set of 5,000 randomly sampled validation samples. The training lasted for 4,100 epochs, with a batch size of 1,024, over a span of about six days. Supplementary Figure S5 illustrates the training and validation curves. We selected the latest checkpoint as the final weight configuration for our Box2Mask module.

Once we trained the Box2Mask module, we allocated one GPU and 8 CPUs for inference evaluation, as well as for the compared methods, to ensure fair comparisons of inference time and resource usage. For the inference phase, we first cropped the ROI images from the promptable bounding boxes, then normalized them using a series of candidate minimum and maximum parameters. The minimum parameters are determined using the 5th to 40th percentiles (in steps of 1), while the maximum parameters are determined using the 90th to 95th percentiles (in steps of 0.5). These parameters are then combined to standardize the ROI images, resulting in candidate normalized ROI images. Subsequently, the Box2Mask module is employed to predict the foreground. The final standardization parameters, $v_{min}$ and $v_{max}$, are determined based on the 0.5th and 99.5th percentile values of the pixel locations predicted as foreground. These parameters are then used to standardize the ROI images for final prediction.

For the PropMask module, throughout the training process, we randomly selected 10,000 tasks per epoch. Each task consists of the guiding slice and four randomly sampled adjacent slices. Evaluations were conducted every 20 epochs using a set of 5,000 randomly selected validation tasks. The training extended over 4,500 epochs, with a batch size of 160, lasting approximately seven days. Supplementary Figure S8 displays the training and validation loss curves. We chose the most recent checkpoint as the final weight configuration for our PropMask module.

\subsection{Loss function}
The Box2Mask module employs deep supervision, enabling predictions at six distinct stages, denoted as $\{\mathbf{P}_{1},\cdots,\mathbf{P}_{S}\}_{S=6}$. Each prediction, $\mathbf{P}_s$, is activated by the Sigmoid function, outputting a 2D representation where values between $[0.0, 1.0]$ indicate the probability of each pixel being part of the foreground.

To align with these six stages of predictions, the foreground ground truth is rescaled to corresponding resolutions, represented as $\{\mathbf{M}_{1},\cdots,\mathbf{M}_{S}\}_{S=6}$, where each $\mathbf{M}_s$ is binary with 1 indicating the foreground.

The loss function is calculated by applying soft dice loss at each stage and then computing the average of these losses. This approach ensures that the module is trained effectively across all resolutions, enhancing its predictive accuracy and reliability. The overall loss function is expressed as:
\begin{align}
    L_{\text{Box2Mask}} = \frac{1}{S}\sum_{s=1}^S\left (1.0 - \frac{2\times\sum_{i=1}^{W_s}\sum_{j=1}^{H_s} \mathbf{P}_{s,i,j} \mathbf{M}_{s,i,j}}{\sum_{i=1}^{W_s}\sum_{j=1}^{H_s} \mathbf{P}_{i,j}^2 + \sum_{i=1}^{W_s}\sum_{j=1}^{H_s} \mathbf{M}_{i,j}^2} \right ),
\end{align}
where $W_s$ and $H_s$ denote the width and height of the resolution at the $s$th stage. The resulting loss value ranges from 0.0 to 1.0.

\subsection{Evaluation metrics}
We evaluated the model from two aspects. For segmentation performance evaluation, we used the Dice Similarity Coefficient (DSC) to evaluate the segmentation results. DSC is a set similarity metric commonly used to calculate the similarity between two samples, with a value range [0.0,1.0]. The more accurate the segmentation result, the larger the DSC, and the closer it is to 1.
\begin{equation}
DSC = \frac{2 |A \cap B|}{|A| + |B|} 
\end{equation} 
where $A$ represents the ground truth mask, $B$ represents the predicted mask, and $| \cdot |$ denotes the cardinality of a set (i.e., the number of elements in the set).

For efficiency evaluation, we used the inference time by recording the time from reading samples to writing final segmentation results. The lower the inference time, the more efficient the model is.

\subsection{Object Complexity Evaluation}
\label{sec:object_complexity_metrics}
To quantify the complexity and irregularity of segmentation targets, we employed three metrics:

Box Ratio: This metric evaluates the similarity between the target mask and its tight bounding box, defined as:
\begin{equation}
\text{BoxRatio}(M)=\frac{|M|}{|\text{Box}(M)|}
\end{equation} 
where $\text{Box}(M)$ is the tightest bounding box around mask $M$, and $|\cdot|$ represents the area measured in pixels.

Convex Ratio: This measure quantifies how convex the target mask is, expressed as:
\begin{equation}
\text{ConvexRatio}(M)=\frac{|M|}{|\text{ConvexHull}(M)|}
\end{equation} 
where $\text{ConvexHull}(M)$ is the convex hull of mask $M$.

Inverse Rotational Inertia (IRI): This metric assesses how spread out the area of the target mask is, calculated as:
\begin{equation}
\text{IRI}(M) = \frac{0.75|M|}{(0.8\pi \text{RI}(M))^{\frac{5}{3}}}
\end{equation} 
where the rotational inertia of $M$ relative to its centroid $c_{M}$ is $\text{RI}(M)=\sum_{x\in\textit{M}} \Vert x - c_{M} \Vert_2^2$, with $x$ being the coordinate of each pixel in the mask, and $c_M$ being the centroid coordinate.

These metrics allow us to objectively assess the complexity of various segmentation targets, providing a basis for evaluating model performance across objects of varying irregularity.

\subsection{Statistics and reproducibility}
Sample sizes and the number of datasets were determined based on the availability of all publicly accessible datasets that we could download and proces. No statistical method was used to pre-determine the sample sizes or the number of datasets.

For performance evaluation and statistical analysis:
\begin{itemize}
    \item DSC calculations were performed for every object type in all datasets, with the DSC of a dataset obtained by averaging the DSCs of its constituent objects.
    \item Performance comparisons across multiple experimental groups in ablation studies were conducted using a one-way ANOVA test. A $P$-value greater than 0.05 indicated no significant difference in performance across the groups, suggesting stability in the experimental results.
    \item Differences in inference speed and interaction time between models were assessed using the Wilcoxon rank-sum test. A $P$-value less than 0.05 indicated significant differences.
    \item We used paired t-test to compare the differences between the improvements of different models.
    \item The relationship between model performance improvements and object irregularity was modeled using a linear model. A negative correlation coefficient ($r$) indicates that greater irregularity is associated with more pronounced model improvements.
\end{itemize}

We used R (version 4.1.3) for results analysis and statistical analyses, and Python (version 3.7.10) for model construction, training, and inference. To ensure reproducibility, we have detailed our methodology in the supplementary materials, which include data collection and processing, module details, definitions of loss functions, experimental specifics, and evaluation metrics. These details can be found in Supplementary Figures S1–S10, Supplementary Tables S1–S10, and Supplementary Text S1. All procedures adhered to good clinical practice and data privacy regulations.

\section*{Data availability statements}
All datasets referenced in this study are publicly available. Supplementary Table S1 provides the download links. The source code and supporting materials are available on GitHub\footnote{https://github.com/czifan/PAM}. All R packages employed in this study can be found on CRAN\footnote{https://cran.r-project.org/web/packages/available\_packages\_by\_name.html} or Bioconductor\footnote{https://www.bioconductor.org/}.

\clearpage
\newpage
\bibliographystyle{naturemag}
\bibliography{sn-bibliography}

\end{document}